\documentclass[11pt,a4paper]{article}
\usepackage[hyperref]{emnlp2020}
\usepackage{times}
\usepackage{latexsym}

\usepackage{url}
\usepackage[whole]{bxcjkjatype}

\usepackage{graphicx}
\usepackage{array}
\usepackage{multirow}
\usepackage{multicol}
\usepackage{comment}
\usepackage{amsfonts}
\usepackage{caption}
\usepackage{subcaption}
\usepackage{amstext}
\usepackage{amsmath}
\usepackage{booktabs}
\usepackage{microtype}
\usepackage{appendix}
\usepackage{wrapfig}
\usepackage{xcolor}

\newcommand{\K}{\texttt{k}}
\newcommand{\waitk}{\texttt{wait-k} }
\newcommand{\wait}{\texttt{wait-} }

\newcommand{\winM}{$^{\dagger}$} 

\newcommand{\Tab}[1]{Table~\ref{tab:#1}}

\newcommand{\unidir}[2]{{#1$\rightarrow$#2}}

\usepackage{pifont}

\usepackage{microtype}

\aclfinalcopy 
\def\aclpaperid{26} 

\title{Towards Multimodal Simultaneous Neural Machine Translation}

\author{Aizhan Imankulova\Thanks{ These authors contributed equally to this paper}\quad
Masahiro Kaneko\footnotemark[1]\quad Tosho Hirasawa\footnotemark[1]\quad Mamoru Komachi\\
Tokyo Metropolitan University\\
6-6 Asahigaoka, Hino, Tokyo 191-0065, Japan\\
{\tt \{imankulova-aizhan, kaneko-masahiro, hirasawa-tosho\}@ed.tmu.ac.jp}\\
{\tt komachi@tmu.ac.jp}}

\date{}

\begin{document}
\maketitle
\begin{abstract}
\label{sec:abst}

Simultaneous translation involves translating a sentence before the speaker's utterance is completed in order to realize real-time understanding in multiple languages.
This task is significantly more challenging than the general full sentence translation because of the shortage of input information during decoding.
To alleviate this shortage, we propose multimodal simultaneous neural machine translation (MSNMT), which leverages visual information as an additional modality. 
Our experiments with the Multi30k dataset showed that MSNMT significantly outperforms its text-only counterpart in more timely translation situations with low latency.
Furthermore, we verified the importance of visual information during decoding by performing an adversarial evaluation of MSNMT, where we studied how models behaved with incongruent input modality and analyzed the effect of different word order between source and target languages. 

\end{abstract}

\section{Introduction}
\label{sec:intro}

Simultaneous translation is a natural language processing (NLP) task in which translation begins before receiving the whole source sentence. 
It is widely used in international summits and conferences where real-time comprehension is one of the essential aspects. 
Simultaneous translation is already a difficult task for human interpreters because the message must be understood and translated while the input sentence is still incomplete, especially for language pairs with different word orders (e.g.\ SVO-SOV)~\cite{seeber2015}.
Consequently, simultaneous translation is more challenging for machines. 
Previous works attempt to solve this task by predicting the sentence-final verb~\cite{grissom2014don}, or predicting unseen syntactic constituents~\cite{oda2015syntax}. 
Given the difficulty of predicting future inputs based on existing limited inputs, \newcite{ma-etal-2019-stacl} proposed a simple simultaneous neural machine translation (SNMT) approach $\waitk$ which generates the target sentence concurrently with the source sentence, but always $\K$ tokens behind, satisfying low latency requirements.

However, previous approaches solve the given task by solely using the text modality, which may be insufficient to produce a reliable translation.
Simultaneous interpreters often consider various additional information sources such as visual clues or acoustic data while translating~\cite{seeber2015}. 
Therefore, we hypothesize that using supplementary information, such as visual clues, can also be beneficial for simultaneous machine translation. 

To this end, we propose Multimodal Simultaneous Neural Machine Translation ({\bf MSNMT}) that supplements the incomplete textual modality with visual information, in the form of an image. It will predict still missing information to improve translation quality during the decoding process.
Our approach can be applied in various situations where visual information is related to the content of speech such as presentations with slides (e.g.\ TED Talks\footnote{https://interactio.io/}) and news video broadcasts\footnote{https://www.a.nhk-g.co.jp/bilingual-english/broadcast/nhk/index.html}.
Our experiments show that the proposed MSNMT method achieves higher translation accuracy than the SNMT model that does not use images by leveraging image information.
To the best of our knowledge, we are the first to propose the incorporation of visual information to solve the problem of incomplete text information in SNMT.

The main contributions of our research are as follows. We propose to combine multimodal and simultaneous NMT, therefore, discovering cases where such multimodal signals are beneficial for the end-task. Our MSNMT approach brings significant improvement in simultaneous translation quality by enriching incomplete text input information using visual clues.
As a result of a thorough analysis, we conclude that the proposed method is able to predict tokens that have not appeared yet for source-target language pairs with different word order (e.g.\ \unidir{English}{Japanese}).
By providing an adversarial evaluation, we showed that the models indeed utilize visual information.

\section{Related Work}
\label{sec:related}

For simultaneous translation, it is crucial to predict the words that have not appeared yet.
For example, it is important to distinguish nouns in SVO-SOV translation and verbs in SOV-SVO translation~\cite{ma-etal-2019-stacl}.
SNMT can be realized with two types of policy: fixed and adaptive policies~\cite{Zheng_2019}.
Adaptive policy decides whether to wait for another source word or emit a target word in one model.
Previous models with adaptive policies include explicit prediction of the sentence-final verb~\cite{grissom2014don,Matsubara2001SIMULTANEOUSJI} and unseen syntactic constituents~\cite{oda2015syntax}. 
Most dynamic models with adaptive policies~\cite{gu2017learning,dalvi-etal-2018-incremental,arivazhagan-etal-2019-monotonic,zheng2019simpler,zheng-etal-2019-speculative,zheng-etal-2020-opportunistic} have the advantage of exploiting input text information as effectively as possible due to the lack of such information in the first place.
Meanwhile, \newcite{ma-etal-2019-stacl} proposed a simple $\waitk$ method with fixed policy, which generates the target sentence only from the source sentence that is delayed by $\K$ tokens.
However, their model for simultaneous translation relies only on the source sentence. 
In this research, we concentrate on the $\waitk$ approach with fixed policy, so that the amount of input textual context can be controlled to analyze better whether multimodality is effective in SNMT.

Multimodal NMT (MNMT) for full-sentence machine translation has been developed to enrich text modality by using visual information~\cite{hitschler2016multimodal,specia2016shared,elliott2017imagination}.
While the improvement brought by visual features is moderate, their usefulness is proven by
\newcite{caglayan-etal-2019-probing}.
They showed that MNMT models are able to capture visual clues under limited textual context, where source sentences are synthetically degraded by color deprivation, entity masking, and progressive masking.
However, they use an artificial setting where they deliberately deprive the models of source-side textual context by masking. However, our research has discovered an actual end-task and has shown the effectiveness of using multimodal data for it. 
Compared with the entity masking experiments \cite{caglayan-etal-2019-probing}, where they use a model exposed to only $\K$ words, our model starts by waiting for the first $\K$ source words and then generates each target word after receiving every new source token, eventually seeing all input text.

In MNMT, visual features are incorporated into standard machine translation in many ways. 
Doubly-attentive models are used to capture the textual and visual context vectors independently and then combine these context vectors in a concatenation manner~\cite{calixto2017doubly} or hierarchical manner~\cite{libovicky-helcl-2017-attention}. 
Some studies use visual features in a multitask learning scenario~\cite{elliott2017imagination,zhou2018vagnmt}.
Also, recent work on MNMT has partly addressed lexical ambiguity by using visual information~\cite{elliott2017findings,lala2018multimodal,gella2019cross} showing that using textual context with visual features outperform unimodal models.

In our study, visual features are extracted using image processing techniques and then integrated into an SNMT model as additional information, which is supposed to be useful to predict missing words in a simultaneous translation scenario.
To the best of our knowledge, this is the first work that incorporates external knowledge into an SNMT model.

\section{Multimodal Simultaneous Neural Machine Translation Architecture}
\label{sec:proposed}

Our main goal is to investigate if image information would bring improvement on SNMT. As a result, two tasks could benefit from each other by combining them. 

In this section, we describe our MSNMT model, which is composed by combining an SNMT framework $\waitk$~\cite{ma-etal-2019-stacl} and a multimodal model~\cite{libovicky-helcl-2017-attention}.
We base our model on the RNN architecture, which is widely used in MNMT research~\cite{libovicky-helcl-2017-attention,caglayan-etal-2017-lium,elliott2017imagination,zhou2018vagnmt,hirasawa2019multimodal}.
The model takes a sentence and its corresponding image as inputs.
The decoder of the MSNMT model outputs the target language sentence in a simultaneous and multimodal manner by attaching attention not only to the source sentence but also to the image related to the source sentence.\footnote{Our code is publicly available at: \url{https://github.com/toshohirasawa/mst}. We fixed our code based on the comments of Ozan Caglayan.}

\subsection{Simultaneous Translation}
\label{sec:snmt}

We first briefly review standard NMT to set up the notations. 
The encoder of standard NMT model always takes the whole input sequence ${\boldsymbol{\rm X}} = (x_1, ..., x_n)$ of length $n$ where each $x_i$ is a word embedding and produces source hidden states ${\boldsymbol{\rm H}} = (h_1, ..., h_n)$.
The decoder predicts the next output token $y_t$ using ${\boldsymbol{\rm H}}$ and previously generated tokens, denoted ${\boldsymbol{\rm Y}}_{<t} = (y_1,...,y_{t-1})$.
The final output is calculated using the following equation:
\begin{equation}
 p({\boldsymbol{\rm Y}}|{\boldsymbol{\rm X}}) = \prod_{t=1}^{|\boldsymbol{\rm Y}|} p(y_t| {\boldsymbol{\rm X}}, y_{<t})
\end{equation}

Different from standard neural translation, in which each $y_i$ is predicted using the entire source sentence ${\boldsymbol{\rm X}}$, the simultaneous translation requires the model to translate concurrently with the growing source sentence.
We incorporate the $\waitk$ approach~\cite{ma-etal-2019-stacl} for our simultaneous translation model. 
Instead of waiting for the whole sentence before translating, this model waits for only the first $\K$ tokens and starts to generate each target tokens after taking every new source token one by one. 
It stops taking new input tokens once the whole input sentence is on board. 
For example, if $\K = 3$, the first target token is predicted using the first 3 source tokens, and the second target token using the first 4 source tokens.
The $\waitk$ decoding probability $p_{\waitk}$ is:
\begin{equation}
 p_{\waitk}({\boldsymbol{\rm Y}}|{\boldsymbol{\rm X}}) = \prod_{t=1}^{|\boldsymbol{\rm Y}|} p(y_t| \boldsymbol{\rm X}_{\leq g(t)}, y_{<t})
\end{equation}
where $g(t)$ is the $\waitk$ policy function which decides how much input text to read and translate, $\boldsymbol{\rm X}_{\leq g(t)} = (x_1, ..., x_{g(t)})$ and $g(t)$ is $0 \leq t \leq n$.
$g(t)$ is defined as follows:
\begin{equation}
 g(t) = \min\{k + t -1, n\}
\end{equation}
When $k + t - 1$ is over source length $n$, $g(t)$ is fixed to $n$, which means the remaining target tokens (including current step) are generated using the full source sentence.
For full sentence translation, $g(t)$ is constant $g(t) = n$.

\subsection{Multimodal Translation}
\label{sec:mmt}

We use a hierarchical attention combination technique~\cite{libovicky-helcl-2017-attention} to incorporate visual and textual features into an MNMT model. 
This model calculates the independent context vectors from the textual features $\boldsymbol{h}^{\rm txt}=(h_1^{\rm txt},...,h_n^{\rm txt})$ and the visual features $\boldsymbol{h}^{\rm img}=(h_1^{\rm img},...,h_m^{\rm img})$, which are extracted by the textual encoder and the image processing model, respectively. It then combines the resulting two vectors using a second attention mechanism, which helps to perform simultaneous translation taking into account visual information.

Specifically, we compute the context vectors $c_{i}^{\rm f}$ for each image ($\rm f=\rm img$) and text ($\rm f=\rm txt$) modality independently using the following equations:
\begin{eqnarray}
 e_{i,j}^{\rm f} & = & \Omega^{\rm f} (s_i, h_{j}^{\rm f}) \\
 \alpha_{i,j}^{\rm f} & = & \frac{{\rm exp}(e_{i,j}^{\rm f})}{\sum_{l=1}^{|\boldsymbol{\rm h}^{\rm f}|}{\rm exp}(e_{i,l}^{\rm f})} \\
 c_{i}^{\rm f} & = & \sum_{j=1}^{|\boldsymbol{\rm h}^{\rm f}|} \alpha_{i,j}^{\rm f} h_{j}^{\rm f} \label{eq:tc}
\end{eqnarray}
where $\Omega^{\rm f}$ is a feedforward network for each modality $\rm f$; $s_i$ is $i$-th decoder hidden state.

We project these image and text context vectors into a common space and compute another distribution over the projected context vectors and their corresponding weighted average using the second attention:
\begin{eqnarray}
 \tilde{e}_{i}^{\rm f} & = & \Psi (s_i, c_{i}^{\rm f}) \\
 \beta_{i}^{\rm f} & = & \frac{{\rm exp}(\tilde{e}_{i}^{\rm f})}{\sum_{\rm r \in \{\rm img, \rm txt\}}{\rm exp}(\tilde{e}_{i}^{\rm r})} \label{eq:b}\\
 \tilde{c}_i & = & \sum_{\rm r \in \{\rm img, \rm txt\}} \beta_{i}^{\rm r} W^{\rm r} c_{i}^{\rm r} \label{eq:hc}
\end{eqnarray}
where $\Psi$ is a feedforward network.
Equation \ref{eq:b} calculates the second attention to combine the image and text vectors.
$W^{\rm r}$ is a weight matrix used to compute the context vector $\tilde{c}_i$ calculated from image and text features.
The final hypothesis ${\boldsymbol{\rm Y}}$ has the probability:
\begin{equation}
 p_{\rm mnmt}({\boldsymbol{\rm Y}}|{\boldsymbol{\rm X}}, {\boldsymbol{\rm Z}}) = \prod_{t=1}^{|\boldsymbol{\rm Y}|} p(y_t| \boldsymbol{\rm X}, \boldsymbol{\rm Z}, y_{<t})
\end{equation}
where ${\boldsymbol{\rm Z}}$ represents input image features.

\subsection{Multimodal Simultaneous Neural Machine Translation}
\label{sec:msnmt}

In this subsection, we describe the structure of the MSNMT model, which is a combination of the models described in Sections \ref{sec:snmt} and \ref{sec:mmt}.
The method for calculating the image context vector is the same as for MNMT; however, the text context vector (Equation \ref{eq:tc}) for the $t$-th step is calculated as follows:
\begin{equation}
 \hat{c}_i^{\rm txt} = \sum_{j=1}^{g(t)} \alpha_{i,j}^{\rm txt} h_{j}^{\rm txt}
\end{equation}
Thus $\hat{c}_i^{\rm txt}$ is calculated from the input text prefix determined by $\waitk$ policy function $g(t)$.
Then we apply the second attention to $\hat{c}_i^{\rm txt}$ and $c_i^{\rm img}$ in order to calculate $\tilde{c}_i$ (Equation \ref{eq:hc}).

The decoding probability becomes as follows:
\begin{equation}
 p_{\rm msnmt}({\boldsymbol{\rm Y}}|{\boldsymbol{\rm X}}, {\boldsymbol{\rm Z}}) = \prod_{t=1}^{|\boldsymbol{\rm Y}|} p(y_t| \boldsymbol{\rm X}_{\leq g(t)}, \boldsymbol{\rm Z}, y_{<t})
\end{equation}

\section{Experimental Setup}
\label{sec:setup}

\subsection{Dataset}
\label{sec:dataset}

We experiment with our model in four translation directions consisting of 5 languages: English (En), German (De), French (Fr), Czech (Cs), and Japanese (Ja). All language pairs include En on the source side.

We used the train, development, and test sets from the Multi30k~\cite{W16-3210} dataset published in the WMT16 Shared Task, which is a benchmark dataset generally used in MNMT research~\cite{libovicky-helcl-2017-attention,caglayan-etal-2019-probing,elliott2017imagination,zhou2018vagnmt,hirasawa2019multimodal} for \unidir{En}{De}, \unidir{En}{Fr} and \unidir{En}{Cs}.

\newcite{nakayama2020visually} released F30kEnt-JP dataset\footnote{\url{https://github.com/nlab-mpg/Flickr30kEnt-JP}} which contains Japanese translations of first two original English captions for each image of the Flickr30k Entities dataset \cite{plummer2017flickr30k}.
They follow the same annotation rules as the Flickr30k Entities dataset using exactly the same tags with entity types and IDs.
We preprocessed this data as follows: 
1) The parallel \unidir{En}{Ja} data was created by taking alignment using corresponding IDs assigned to each Japanese translation entity with the IDs of Flickr30k entities.\footnote{We used the second translations due to some empty translations of the first captions.}
2) The created parallel data was aligned with its corresponding images using text files named $(image\_id).txt$ corresponding to each image in Flickr30k.
3) Finally, the created multimodal data was split to train, dev, and test following data splits of Multi30k using the same Multi30k image IDs.
Note that the English side of \unidir{En}{Ja} parallel data extracted from F30kEnt-JP and English side of Multi30k data are thought to be somewhat comparable but not strictly the same while their corresponding images are the same. 

Data split for all language pairs were as follows: training set, 29,000 sentence pairs, development set, 1,014 sentence pairs, and 1,000 sentence pairs for the test set.
This dataset's average sentence length is 12-13 tokens for En, De, Fr, Cs and 20 tokens for Ja.

We limit the vocabulary size of the source and the target languages after concatenating them to 10,000 sub-words \cite{sennrich-haddow-birch:2016:P16-12}. 
All sentences are preprocessed with lower-casing, tokenizing, and normalizing the punctuation using the Moses script\footnote{We applied preprocessing using task1-tokenize.sh from https://github.com/multi30k/dataset.}. 
To tokenize Japanese sentences, we used MeCab\footnote{\url{http://taku910.github.io/mecab}, version 0.996.} with the IPA dictionary.

Visual features are extracted using pre-trained ResNet~\cite{he2016deep}. Technically, we encode all images in Multi30k with ResNet-50 and pick out the hidden state in the pool5 layer as a 2,048-dimension visual feature.

\begin{table*}[t]
\centering
\scalebox{.8}{
\begin{tabular}{crrrrrrrr}
\toprule
\multirow{2}{*}{\waitk} & \multicolumn{2}{c}{\unidir{En}{De}} & \multicolumn{2}{c}{\unidir{En}{Fr}} & \multicolumn{2}{c}{\unidir{En}{Cs}} & \multicolumn{2}{c}{\unidir{En}{Ja}} \\
\cmidrule(r){2-3} \cmidrule(r){4-5} \cmidrule(r){6-7} \cmidrule(r){8-9}
    & \multicolumn{1}{c}{S}     & \multicolumn{1}{c}{M}    & \multicolumn{1}{c}{S}     & \multicolumn{1}{c}{M}    & \multicolumn{1}{c}{S}     & \multicolumn{1}{c}{M}   & \multicolumn{1}{c}{S}     & \multicolumn{1}{c}{M}    \\
 \midrule
1    & 19.18 & {\winM}\bf{19.90} & 31.23 & {\winM}\bf{32.49} & 7.78    & {\winM}\bf{9.07} & 21.95  &  {\winM}\bf{23.45}    \\
3    & 28.22 & {\winM}\bf{28.75} & 43.85 & \bf{43.99}        & 18.91   & {\winM}\bf{19.39} & 27.35  &  {\winM}\bf{27.74}    \\
5    & 30.38 & {\winM}\bf{31.48} & 48.01 & {\winM}\bf{48.40} & 23.35   & \bf{23.50}       & 31.71  &  \bf{31.72}    \\
7    & 31.72 & \bf{32.14}          & 50.14 & \bf{50.16}          & 25.65   & \bf{25.83}       & 33.70  &  \bf{33.93}    \\
Full & 34.64 & \bf{34.84}        & 53.55 & \bf{53.78}         & \bf{27.22}  & 26.85        & \bf{35.93} &  35.62    \\
\bottomrule
\end{tabular}
}
\caption{BLEU scores of SNMT (S) and MSNMT (M) models for four translation directions on test set. Results are the average of four runs. {\bf Bold} indicates the best BLEU score for each $\waitk$ for each translation direction. ``{\winM}'' indicates statistical significance of the improvement over SNMT.}
\label{tab:direct}
\end{table*}

\begin{figure*}[t!]
\centering
    \begin{subfigure}[c]{0.4\textwidth}
        \centering
        \includegraphics[width=\linewidth]{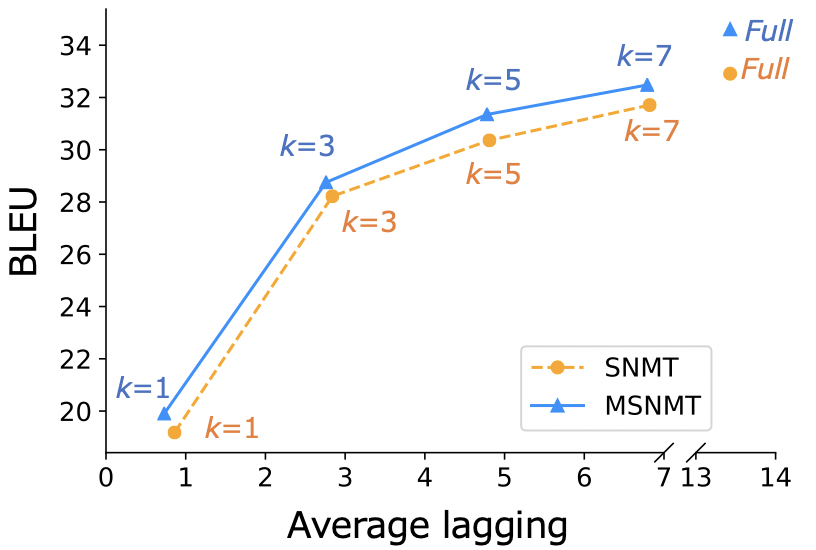}
        \caption{\unidir{En}{De}}
        \label{fig:ende}
    \end{subfigure}
    \qquad\qquad
    \begin{subfigure}[c]{0.4\textwidth}
        \centering
        \includegraphics[width=\linewidth]{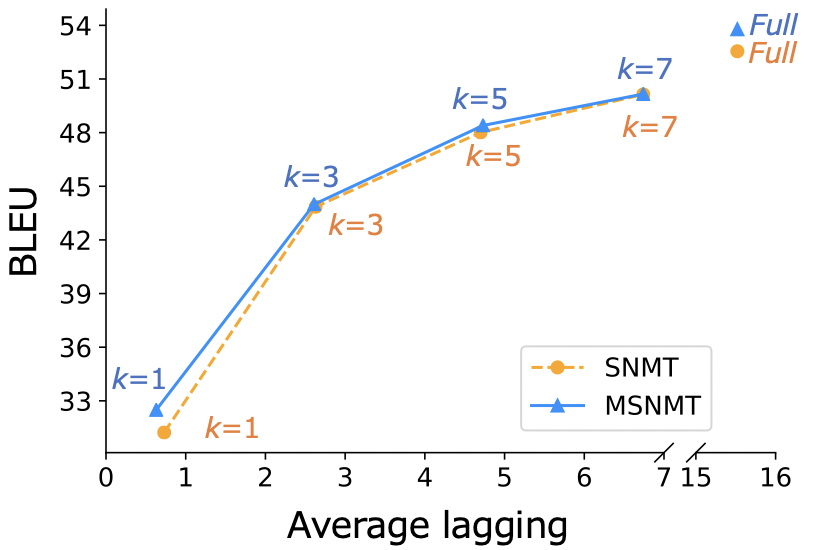}
        \caption{\unidir{En}{Fr}}
        \label{fig:enfr}
    \end{subfigure}
    \qquad
    \begin{subfigure}[c]{0.4\textwidth}
        \centering
        \includegraphics[width=\linewidth]{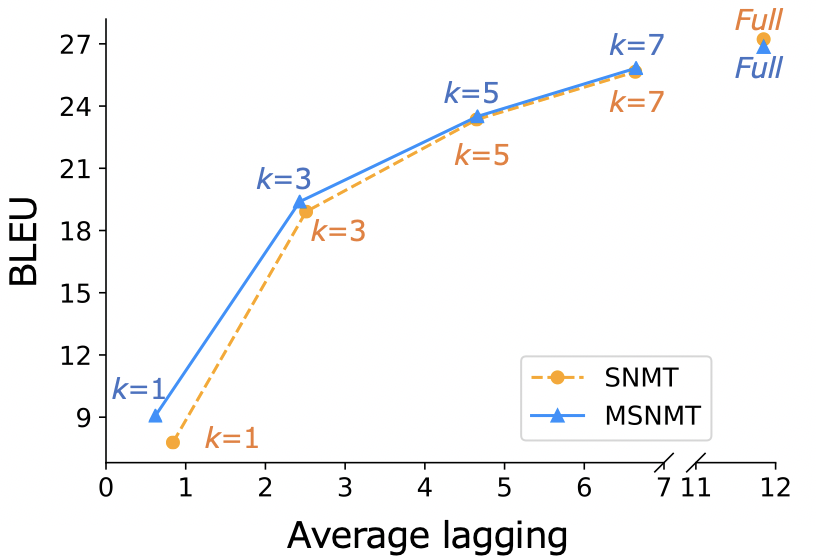}
        \caption{\unidir{En}{Cs}}
        \label{fig:encs}
    \end{subfigure}
    \qquad\qquad
    \begin{subfigure}[c]{0.4\textwidth}
        \centering
        \includegraphics[width=\linewidth]{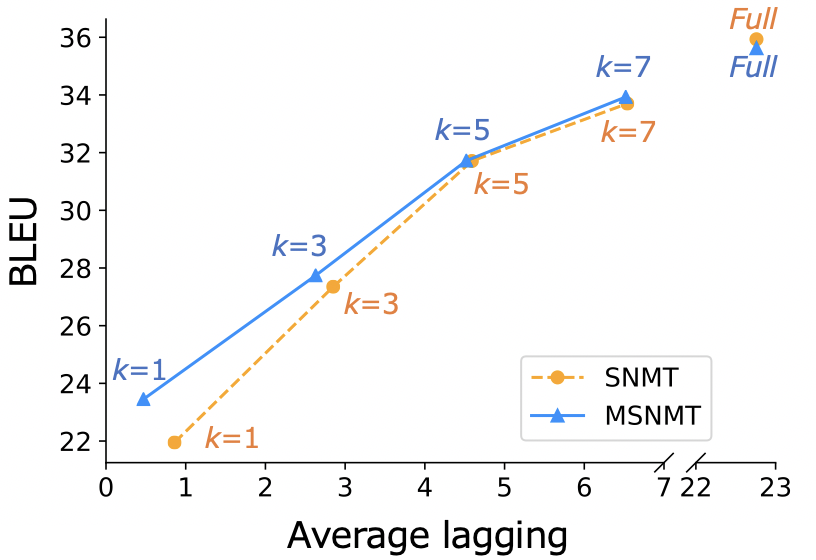}
        \caption{\unidir{En}{Ja}}
        \label{fig:enja}
    \end{subfigure}\hfill
    \caption{Average Lagging scores. Results are the average of four runs.}\label{fig:al}
\end{figure*}

\subsection{Systems}
\label{sec:systems}

We compare the following models:
{\bf 1. SNMT:} We use only text modality for training data as a baseline for each $\waitk$ model.
{\bf 2. MSNMT:} We use image modality along with text modality for a training data for each $\waitk$ model.

To train the above models, we utilize attention NMT~\cite{bahdanau} with a 2-layer unidirectional GRU encoder and a 2-layer conditional GRU decoder.
We use the open-source implementation of the \texttt{nmtpytorch} toolkit v3.0.0~\cite{caglayan2017nmtpy}.
We first pre-train the MSNMT model for each $\K$ until convergence using only text data and use zeros for visual features. 
Then we continue training MSNMT on multimodal data for each $\K$.
We employ early-stopping: the training was stopped when the BLEU score did not increase on the development set for 10 epochs for MSNMT pre-training, 5 epochs for MSNMT fine-tuning, and 15 epochs for SNMT training.

In order to keep our experiments as pure as possible, we will not use additional data or other types of models. It will allow us to control the amount of input textual context, so we can easily analyze the relationship between the amount of textual and visual information.

\begin{table*}[t]
\centering
 \scalebox{.8}{
\begin{tabular}{crrrrrrrr}
\toprule
\multirow{2}{*}{\waitk} & \multicolumn{2}{c}{\unidir{En}{De}} & \multicolumn{2}{c}{\unidir{En}{Fr}} & \multicolumn{2}{c}{\unidir{En}{Cs}} & \multicolumn{2}{c}{\unidir{En}{Ja}} \\
\cmidrule(r){2-3} \cmidrule(r){4-5} \cmidrule(r){6-7} \cmidrule(r){8-9}
    & \multicolumn{1}{c}{C}     & \multicolumn{1}{c}{I}    & \multicolumn{1}{c}{C}     & \multicolumn{1}{c}{I}    & \multicolumn{1}{c}{C}     & \multicolumn{1}{c}{I}   & \multicolumn{1}{c}{C}     & \multicolumn{1}{c}{I}    \\
 \midrule
1    & {\winM}\textbf{19.90} & 8.19  & {\winM}\textbf{32.49} & 18.00 & {\winM}\textbf{9.07}  & 6.83  & {\winM}\textbf{23.45} & 17.57 \\
3    & {\winM}\textbf{28.75} & 26.78 & {\winM}\textbf{43.99} & 42.31 & {\winM}\textbf{19.39} & 18.78 & {\winM}\textbf{27.74} & 24.51 \\
5    & \textbf{31.48} & 31.08 & \textbf{48.40} & 48.19 & {\winM}\textbf{23.50} & 22.81 & {\winM}\textbf{31.72} & 28.57 \\
7    & {\winM}\textbf{32.14} & 32.04 & \textbf{50.16} & 50.15 & {\winM}\textbf{25.83} & 25.09 & {\winM}\textbf{33.93} & 31.03 \\
Full & \textbf{34.84} & 34.40 & {\winM}\textbf{53.78} & 53.10 & \textbf{26.85} & 26.84 & \textbf{35.62} & 35.59 \\
\bottomrule
\end{tabular}
 }
\caption{Image Awareness results on the test set. BLEU scores of MSNMT Congruent (C) and Incongruent (I) settings for four translation directions. Results are the average of four runs. {\bf Bold} indicates the best BLEU score for each $\waitk$ for each translation direction. ``{\winM}'' indicates the statistical significance of the improvement over Incongruent settings.}
\label{tab:adversarial}
\end{table*}

\subsection{Hyperparameters}
\label{sec:appendix_param}

We use the same hyperparameters for SNMT and MSNMT for a fair comparison as follows.
All models have word embeddings of 200 and recurrent layers of dimensionality 400 units with 2way sharing of embeddings in the network. 
We used Adam \cite{kingma:2015:ICLR} with a learning rate of 0.0004.
Decoders were initialized with zeros.
We used a minibatch size of 64 for training and 32 for fine-tuning. 
Rates of dropout applied on source embeddings, source encoder states and pre-softmax activations were 0.4, 0.5, and 0.5, respectively.
We set the max length of the input to 100.
$\waitk$ experiments were conducted for 1, 3, 5, 7, and Full settings.
For MSNMT only hyperparameters, the sampler type was set to approximate, and channels were set to 2048.
The fusion type was set to hierarchical mode.

\subsection{Evaluation}
\label{sec:eval}

We report BLEU scores calculated using Moses' multi-bleu.perl, which is a widely used evaluation metric in MT, on our test sets for each $\waitk$ model.\footnote{Due to space constraints, we show results only for test sets.}
Statistical significance ($p<0.05$) on the difference of BLEU scores was tested by Moses' \textit{bootstrap-hypothesis-difference-significance.pl}. ``Full'' means that the whole input sentence is used as an input for the model to start translating.
All reported results are the average of four runs using four different random seeds.

Additionaly, we use open-sourced Average Lagging (AL) latency metric proposed by \newcite{ma-etal-2019-stacl} to evaluate the latency for SNMT and MSNMT systems.\footnote{\url{https://github.com/SimulTrans-demo/STACL}} 
It calculates the degree of out of sync time with the input, in terms of the number of source tokens as follows:
\begin{equation}
 {\rm AL}_{g}(\boldsymbol{\rm X}, \boldsymbol{\rm Y}) = \frac{1}{\tau_{g}(|\boldsymbol{\rm X}|)} \sum_{t=1}^{\tau_{g}(|\boldsymbol{\rm X}|)} g(t) - \frac{t-1}{r}
\end{equation}
where $r = |\boldsymbol{\rm Y}|/|\boldsymbol{\rm X}|$ is the target-to-source length ratio and $\tau_{g}$ is the decoding step when source sentence finishes:
\begin{equation}
 \tau_{g}(|\boldsymbol{\rm X}|) = \min \{t | g(t) = |\boldsymbol{\rm X}|\}
\end{equation}

\section{Results}
\label{sec:results}

Table \ref{tab:direct} illustrates the BLEU scores of MSNMT and SNMT models on the test set.
MSNMT systems show significant improvements over SNMT systems for all language pairs when input textual information is limited.
Note that the difference of BLEU scores between MSNMT and SNMT becomes larger as the $\K$ gets smaller, especially when the target language is distant from English in terms of word order (e.g.\ Cs and Ja).
On the other hand, the availability of more tokens during the decoding process ($\K$ $\geq$ 5) leads to the text information becoming sufficient in some cases.

Figure \ref{fig:al} shows translation quality against AL for four language directions. 
In all these figures, we observe that, as $\K$ increases, the gap between BLEU scores for MSNMT and SNMT decreases.
We also observe that AL scores are better for MSNMT as $\K$ decreases.
From these results, it can be seen that in terms of latency, the smaller $\K$ is, the more beneficial the visual clues become.

\section{Analysis}
\label{sec:disc}

In this section, we provide a thorough analysis to further investigate the effect of visual data to produce a simultaneous translation by (a) providing adversarial evaluation; and (b) analyzing the impact of different word order for \unidir{En}{Ja} language pair.

\subsection{Adversarial Evaluation}
\label{sec:adv}

In order to determine whether MSNMT systems are aware of the visual context~\cite{elliott2018adversarial}, we perform the adversarial evaluation on the test set.
We present our system with correct visual data with its source sentence (Congruent) as opposed to random visual data as an input (Incongruent)~\cite{elliott2018adversarial}.Therefore, we reversed the order of 1,000 images of the test set, so there will be no overlapping congruent visual data.
Then we reconstruct image features for those images to use as an input. 

Results of image awareness experiments are shown in Table \ref{tab:adversarial}. 
We can see the large difference in BLEU scores between MSNMT congruent (C columns) and incongruent (I columns) settings when $\K$ are small.
This implies that our proposed model utilizes images for translation by learning to extract needed information from visual clues. 
The interesting part is for a full translation, where scores for the incongruent setting are very close to those of the congruent setting.
The reason is that when textual information is enough, visual information becomes not that relevant in some cases.

\subsection{How Source-Target Word Order Affects Translation}
\label{sec:entities}

In $\waitk$ translations, for the \unidir{En}{Ja} language pair with different word orders (SVO vs. SOV), some source tokens should be translated before they are presented to the decoder for grammaticality and fluency purposes.
Hence, the model also needs to handle such cases well apart from the ``usual'' order.
We hypothesized that MSNMT models, given additional visual information, are able to translate such cases better than SNMT models.
Therefore, we investigated how many tokens were correctly translated that are not given as input yet.

\begin{table*}[t]
\centering
\scalebox{.75}{%
\begingroup
\setlength{\tabcolsep}{3pt}
\begin{tabular}{l|cccccccccccccccccc}
\toprule
$t$& 1 & 2 & 3 & 4 & 5 & 6 & 7 & 8 & 9 & 10 & 11 \\
Source & \textcolor{red}{a} & \textcolor{red}{person} & rappelling & \textcolor{brown}{a} & \textcolor{brown}{cliff} & above & \textcolor{blue}{a} & \textcolor{blue}{body} & \textcolor{blue}{of} & \textcolor{blue}{water} & . \\
\midrule
Target, $\K$=3 &&&& \textcolor{blue}{海} & の & 上 & に & ある & \textcolor{brown}{断崖} & を & 降り & て & いる & \textcolor{red}{一} & \textcolor{red}{人} & \textcolor{red}{の} & \textcolor{red}{男性} & 。 \\
Entity count &&&& \ding{51} &&&&& \ding{55} &&&&&&& \ding{55} & \\ 
\bottomrule
\end{tabular}
\endgroup
}%
\caption{Example of \unidir{En}{Ja} translation to count entities that should be translated before introducing it to a model in case of $\wait3$ (see Figure \ref{fig:example_cliff}). A $\waitk$ model starts translating after $\K$ tokens are inputted. Colors represent the same entities. \ding{51} indicates entities that are not presented to the model at timestep $t$ yet and \ding{55} indicates entities that are already seen by the model at timestep $t$.
We count only those entities marked with \ding{51} for \# total entities (Table \ref{tab:entities}).}
\label{tab:entity_count_example}
\end{table*}

\begin{table*}[t]
\centering
\scalebox{.8}{
\begin{tabular}{crrrrr}
\toprule
\multirow{2}{*}{\K} &  \multirow{2}{*}{\# total entities} & \multicolumn{2}{c}{\# correct entities by S}      & \multicolumn{2}{c}{\# correct entities by M}     \\
\cmidrule(r){3-4} \cmidrule(r){5-6}
&  & \waitk  & \texttt{Full} & \waitk & \texttt{Full} \\
 \midrule
1    & 1,343              & 251            & \textbf{716}          & \textbf{270}            & 707          \\
3    & 852               & 229            & \textbf{433}          & \textbf{242}            & 432          \\
5    & 502               & 147            & \textbf{247}          & \textbf{151}            & 243          \\
7    & 320               & 106            & \textbf{160}          & 106            & 159          \\
\bottomrule
\end{tabular}
}
\caption{Number of entities that were correctly translated before being presented to the model by SNMT (S) and MSNMT (M) models with their for each $\K$. Results are the average of four runs.}
\label{tab:entities}
\end{table*}
 
First, we quantitatively analyze how well we can translate entities that are not presented from the source yet but should exist in target sentences.
To align the source and target entities, we use the entities' annotation attached to both the source and target sentences.
Given that annotated entities have the same IDs and tags for both English and Japanese, we can align, calculate, and extract those entities from source and target sentences.
If the index of the first token of the aligned target entity is not given as input at timestep $\K$ yet, we count them for each $\K$ scenario as \# total entities (Table \ref{tab:entities}).
For example, in Table \ref{tab:entity_count_example} a $\wait3$ model should start translating after a token ``rappelling'' is presented to the model.
And if an ID of the entity of ``海 (a body of water)'' is in the target sentences but not in the inputted part yet, we count it as an entity that should be translated before being inputted to the model.
Similarly, an entity of ``断崖 (cliff)'' is already presented to the model at timestep 5, so we do not count those entities.
If the same entity ID appears more than once in one sentence, we exclude those entities due to the impossibility of alignments.
Finally, for each model during decoding, if those entities are included in the model's translation results with a perfect match from pre-calculated \# total entities, we consider them as correctly translated.\footnote{We can not create \# total entities from decoded tokens directly due to unavailability of entity annotations.}

Table \ref{tab:entities} demonstrates the results.
$\K$ column is to determine how many tokens a model waits before starting translating. 
Note that $\K$=Full is not included because all entities are given at the time of translation.
The reason that the total number of entities that were not inputted yet decreases when $\K$ increases (\# total entities column) is that more entities are already available for the model for translation.
$\waitk$ columns show how many entities were correctly translated by $\waitk$ SNMT and MSNMT models from \# total entities for each $\K$ scenario.
Columns \texttt{Full} show upper-bounds of how many entities can be correctly translated if the models were trained with full sentences for entities from each $\K$.
Comparing \texttt{Full} results to $\waitk$ for both SNMT and MSNMT shows that it is hard to correctly translate entities when $\K$ is small. 
Furthermore, comparing $\waitk$ results of SNMT to MSNMT, it can be seen that the smaller value of $\K$, the better MSNMT can handle different source-target word order than SNMT.

\begin{figure}[h!]
\centering
    \begin{subfigure}[b]{0.21\textwidth}
        \centering
        \includegraphics[width=\linewidth]{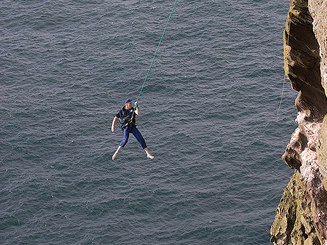}
        \caption{A person rappelling a cliff.}
        \label{fig:example_cliff}
    \end{subfigure}
    \qquad
    \begin{subfigure}[b]{0.21\textwidth}
        \centering
        \includegraphics[width=\linewidth]{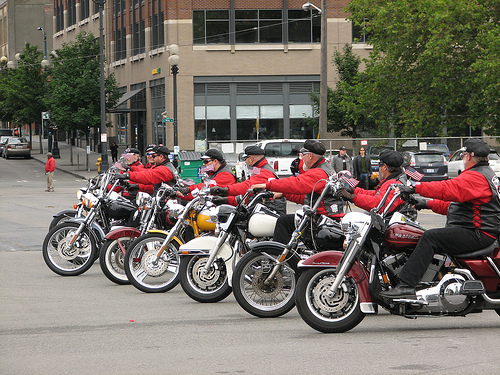}
        \caption{Eight men on motorcycles.}
        \label{fig:example_motor}
    \end{subfigure}
    \hfill
    \caption{Images presented in translation examples (\Tab{examples}).}\label{fig:orig_examples}
\end{figure}

\begin{table*}[h!]
\centering
\scalebox{.75}{%
\begin{tabular}{l|l}
\toprule
Source & \textcolor{red}{a person} rappelling \textcolor{brown}{a cliff} above \textcolor{blue}{a body of water} . \\
Target & \textcolor{blue}{海}\,の\,上\,に\,ある\,\textcolor{brown}{断崖}\,を\,降り\,て\,いる\,\textcolor{red}{一\,人\,の\,男性}\,。 \\
\midrule
S $\wait3$ & \textcolor{red}{誰\,か}\,が\,、\,岩\,の\,上\,で\,\textcolor{brown}{崖}\,を\,登る\,。 (someone climbs a cliff on a rock.)\\
M $\wait3$ & \textcolor{red}{人}\,が\,\textcolor{blue}{海}\,の\,上\,で\,\textcolor{brown}{崖}\,を\,降り\,て\,いる\,。 (a person is rappelling a cliff above the sea.)\\
S \texttt{Full} & \textcolor{red}{人}\,が\,\textcolor{blue}{水域}\,の\,上\,の\,\textcolor{brown}{崖}\,を\,登っ\,て\,いる\,。 (a person is climbing a cliff above a body of water.)\\
M \texttt{Full} & \textcolor{red}{人}\,が\,\textcolor{blue}{水域}\,の\,上\,で\,\textcolor{brown}{崖}\,を\,降り\,て\,いる\,。 (a person is rappelling a cliff above a body of water.)\\
\midrule
\midrule
Source & \textcolor{red}{eight men} on \textcolor{cyan}{motorcycles} dressed in \textcolor{orange}{red and black} are all lined up on \textcolor{blue}{the side of the street} . \\
Target & \textcolor{orange}{赤\,と\,黒}\,の\,服\,を\,着\,た\,\textcolor{cyan}{オートバイ}\,に\,乗っ\,て\,いる\,\textcolor{red}{８\,人\,の\,男性}\,が\,\textcolor{blue}{通り\,の\,脇}\,に\,ずらりと\,並ん\,で\,いる\,。 \\
\midrule
S $\wait3$ & \textcolor{orange}{白い}\,服\,を\,着\,て\,、\,\textcolor{orange}{黒\,と\,黒}\,の\,服\,を\,着\,た\,\textcolor{red}{１\,人\,の\,男性}\,が\,、\,\textcolor{blue}{通り\,の\,脇}\,に\,並ん\,で\,いる\,。 \\
& (a man in white and black and black is standing beside the street.)\\
M $\wait3$ & \textcolor{cyan}{自転車}\,に\,乗っ\,て\,いる\,\textcolor{orange}{赤\,と\,黒}\,の\,服\,を\,着\,た\,\textcolor{red}{８\,人\,の\,男性}\,が\,、\,\textcolor{blue}{通り\,の\,側面}\,に\,ある\,。\\
& (eight men in red and black clothes riding a bicycle are on the side of the street.)\\
S \texttt{Full} & \textcolor{orange}{赤\,と\,黒}\,の\,服\,を\,着\,た\,、\,\textcolor{cyan}{オートバイ}\,に\,乗っ\,た\,\textcolor{red}{２\,人\,の\,男性}\,が\,、\,\textcolor{blue}{通り\,の\,脇}\,で\,並ん\,で\,いる\,。 \\
& (two men on motorcycles, dressed in red and black, line up by the side of the street.)\\
M \texttt{Full} & \textcolor{orange}{赤\,と\,黒}\,の\,服\,を\,着\,た\,、\,\textcolor{cyan}{オートバイ}\,に\,乗っ\,た\,\textcolor{red}{８\,人\,の\,男性}\,が\,、\,\textcolor{blue}{通り\,の\,側面}\,に\,並ん\,で\,いる\,。 \\
& (eight men on motorcycles, dressed in red and black, line the side of the street.)\\
\bottomrule
\end{tabular}
}
\caption{Examples of \unidir{En}{Ja} translations from  test set using SNMT (S) and MSNMT (M) models (also refer to Figure \ref{fig:orig_examples}). In () are shown their English meanings. The same colors indicate the same entity types.}
\label{tab:examples}
\end{table*}

As an example, we sampled sentences and their images from the \unidir{En}{Ja} test set (Figure \ref{fig:orig_examples}) to compare the outputs of our systems.
\Tab{examples} lists their translations generated by SNMT (S) and MSNMT (M) models. 
In the first example, an SNMT model with $\wait3$ could not predict ``海 (sea, a body of water)'' which appears at the end of the source sentence and generated an erroneous ``岩 (rock)'' which is not present neither in source text nor in a corresponding image.
Contrarily, the MSNMT model with $\wait3$ was able to correctly predict ``海 (body of water)'' even before it was inputted by capturing visual information.
When a full sentence is given as an input, MSNMT translated it correctly using more information, unlike SNMT, which translated only from the given text and generated incorrect ``登って (climbing)'' instead of ``降りて (rappelling)''.
Interestingly, in the second example, the MSNMT model with $\wait3$ predicted ``自転車 (bicycles)'' instead of ``オートバイ (motorcycles)'' at the beginning of the sentence, while the SNMT model with $\wait3$ was not able to generate any vehicle entities. 
Also, both MSNMT models with $\wait3$ and \texttt{Full} correctly captured that there were eight men, whilst both SNMT models incorrectly predicted about one and two men.
From these results, we can conclude that visual clues positively impact generated translations where there is still a lack of textual information, especially when we deal with language pairs with different word order.

\section{Conclusion}
\label{sec:conclusion}

In this paper, we proposed a multimodal simultaneous neural machine translation approach, which takes advantage of visual information as an additional modality to compensate for the shortage of input text information in the simultaneous neural machine translation.
We showed that in a $\waitk$ setting, our model significantly outperformed its text-only counterpart in situations where only a few input tokens are available to begin translation.
We showed the importance of the visual information for simultaneous translation, especially in the low latency setup and for a language pair with word-order differences.
We hope that our proposed method can be explored even further for various tasks and datasets.

In this paper, we created a separate model for each value of $\waitk$. However, in future work, we plan to experiment on having a single model for all $\K$ values~\cite{Zheng_2019}.
Furthermore, we acknowledge the importance of investigating MSNMT effects on more realistic data (e.g.\ TED), where the utterance does not necessarily match a shown image while speaking and/or where its context can not be guessed from the shown image.

\section*{Acknowledgments}

We are immensely grateful to Raj Dabre and Rob van der Goot who provided expertise, support, and insightful comments that significantly improved the manuscript. 
We would also like to show our gratitude to Desmond Elliot for valuable feedback and paper discussions.
We want to thank Ozan Caglayan for pointing out critical bugs in our previous implementation.

\bibliography{emnlp2020}
\bibliographystyle{acl_natbib}

\end{document}